\documentclass{article}
\usepackage{spconf,amsmath,epsfig}

\title{Unsupervised Super-Resolution of Satellite Imagery for high fidelity material label transfer}
\name{Arthita Ghosh, Max Ehrlich, Larry Davis, Rama Chellappa}
\address{University of Maryland, College Park, MD, USA}
\begin{document}
\maketitle
\begin{abstract}
    Urban material recognition in remote sensing imagery is a highly relevant, yet extremely challenging problem due to the difficulty of obtaining human annotations, especially on low resolution satellite images. To this end, we propose an unsupervised domain adaptation based approach using adversarial learning. We aim to harvest information from smaller quantities of high resolution data (source domain) and utilize the same to super-resolve low resolution imagery (target domain). This can potentially aid in semantic as well as material label transfer from a richly annotated source to a target domain.
\end{abstract}
\section{Introduction}\label{sec:intro}
Material classification in remote sensing imagery is essential for generating very detailed 3D maps and realistic 3D models of the area. High resolution training data that clearly capture details over vast areas is required in order to generate high fidelity models of man made structures e.g. buildings, roads, bridges from satellite images. Deep networks can be used to automatically and accurately extract these information. However, that calls for supervised training on ground truth annotated by humans which can be challenging to obtain in large numbers. As an example of this scenario we study two datasets - the first (ISPRS Potsdam~\cite{rottensteiner2012isprs}) consists of high resolution images captured at 5cm ground sampling distance. The second is the URBAN3D dataset~\cite{goldberg2018urban}. Satellite Imagery in URBAN3D  is captured at a much higher ground sample distance(50 cm). Lower resolution presents a challenge in extracting detailed features for segmentation/detection etc. as well as capturing exact 3D structures of building roofs for accurate 3D modeling. However, images in the URBAN3D dataset cover a larger area and higher number of structures than other high resolution datasets - which is more conducive to training deep networks. We  combine a larger, low resolution dataset (URBAN3D with building foot print ground truth) with a smaller, high resolution dataset (ISPRS 2D semantic labeling with accurate pixel labels for 6 semantic classes). We use deep domain adaptation techniques for unsupervised super-resolution and semantic label transfer from a small, high resolution, richly annotated source domain to a larger, low resolution target domain.
\section{Related Work}\label{sec:RelatedWork}
\par Material labels for urban structures in synthetic remote sensing imagery can be easily generated using certain modeling software. However, there is a domain gap between real and synthetic satellite imagery that needs to be addressed. An alternative approach is to obtain reliable manual annotations for publicly available high resolution real satellite imagery datasets. Rich, detailed annotations collected in small quantities need to be effectively transferred to larger volumes of unlabeled/partially labeled data  potentially belonging to a different domain. Unsupervised domain adaptation (UDA) has been successfully applied to train deep networks to solve various computer vision tasks, predominantly to transfer knowledge from large volumes of fully annotated synthetic data to lesser quantities of real data ~\cite{sankaranarayanan2017learning,ren2018cross,
    bousmalis2017unsupervised,guo2018learning,atapour2018real,
    chen2018road,wu2018dcan,yang2018real}. Joint optimization of deep networks on inputs from multiple domains has been studied in several recent works~\cite{de2017procedural,hu2018duplex,li2018domain,rebuffi2018efficient,
    volpi2018adversarial,li2018deep,zhang2018fully}. We perform a UDA based joint optimization and super-resolution to close the domain gap between richly and scarcely annotated remote sensing datasets. Super-resolution using adversarial approach has been explored by ~\cite{ledig2017photo,wang2018esrgan}. Unsupervised super-resolution techniques have been proposed by ~\cite{ravi2018adversarial,yuan2018unsupervised,lin2018deep}
\section{Method}\label{sec:method}
\begin{figure*}[tbh]
    \includegraphics[scale=0.32]{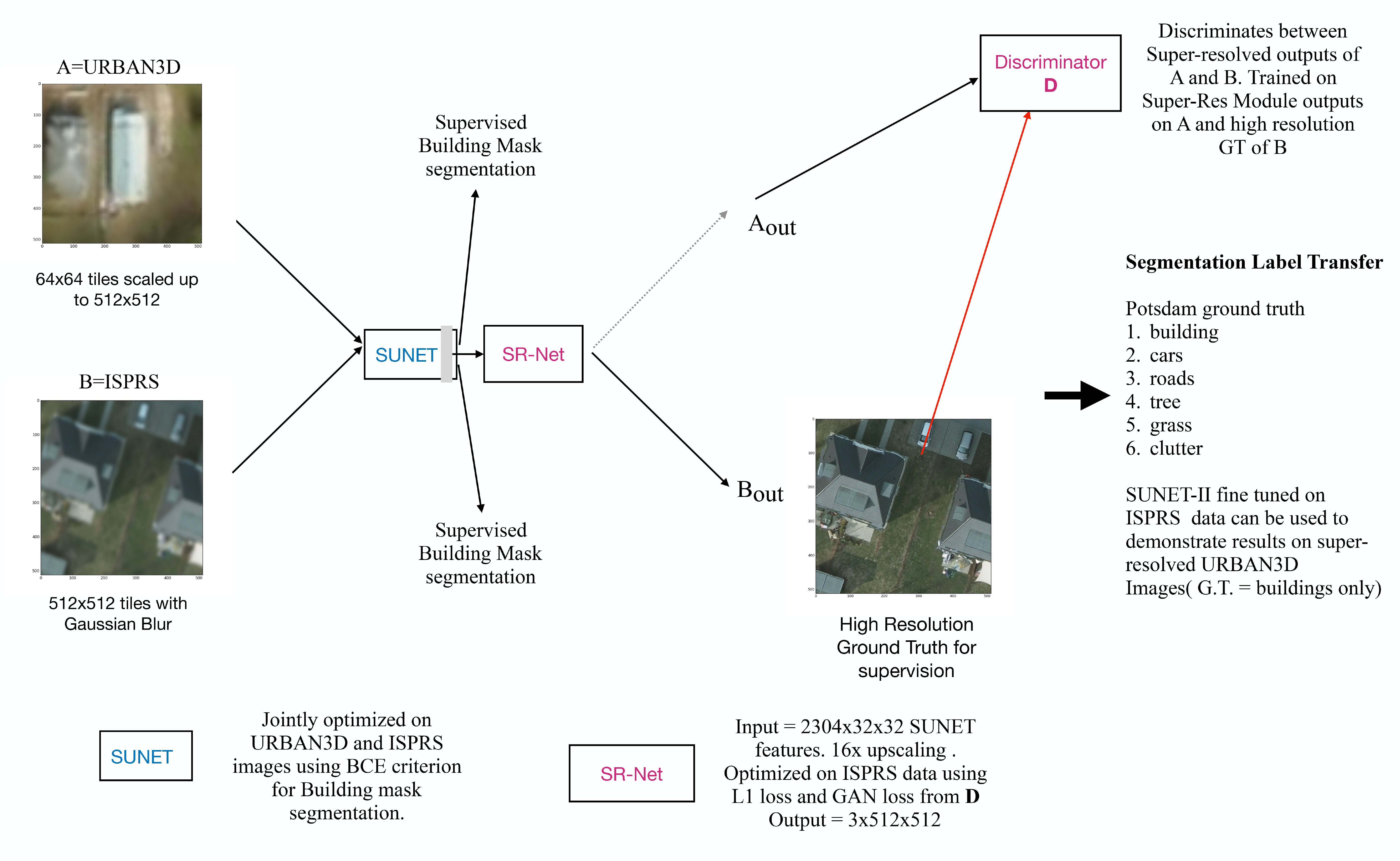}
    \caption{Overview of Unsupervised Domain Adaptation based super-resolution of remote sensing imagery}\label{Fig1}
\end{figure*}
Figure~\ref{Fig1} schematically presents our approach of unsupervised domain adaptation between datasets that consist of high (source) and a low (target) resolution remote sensing imagery respectively. The source domain contains only 38 true orthophoto RGB images from which we randomly crop tiles for training. Target domain contains 2D orthorectified RGB images covering a much larger area over a different region and a much greater number of urban structures.\\

\textbf{Shared latent space} The backbone of the proposed deep domain adaptation technique is a shared common task between the source and target domains used to construct a shared latent space. Both the high resolution ISPRS images and low resolution URBAN3D images have building footprint annotations. We train a building footprint segmentation network jointly on both datasets. We use mixed mini-batches of samples from source and target domains during training. Joint optimization on a shared task is used to align deep features extracted from both kinds of data. The features of the building segmentation network are further used for super-resolution. For building segmentation we use the stacked U-Net (\textbf{SUNET}) architecture as proposed is ~\cite{shah2018stacked} and adapted to satellite imagery in~\cite{ghosh2018stacked}. SUNET consists of 4 blocks of stacked U-Nets containing 2,7,7 and 1 U-Net module(s) respectively. Every module is preceded and succeeded by 1x1 convolutions for feature transformation. Input images pass through 7x7 convolution filters (stride=2) and a residual block. For an input size of 512x512, the output map size is 32x32, This is owing to 2 strided convolutions and 2 average pooling operations. Dilated convolutions are used to keep feature map resolution constant across U-Net modules. The total number of parameters learned is 37.7 million.\\

\textbf{Super-Resolution Module}
The super-resolving deep network (\textbf{SR-Net}) consists of pixel shuffle layers~\cite{shi2016real} that up-sample feature maps to 512x512 high resolution output tiles. The output feature maps of the segmentation network have 2304 channels of dimension 32x32. These are passed through 9x9 convolution filters that reduce the number of channels, a residual block, a 3x3 convolution layer and finally four consecutive up-sampling modules that deploy pixel shuffling to generate 512x512 feature maps. The final convolution layer generates super-resolved three channel output tiles of size 512x512.\\

\textbf{Adversarial Learning}
Direct supervision on SR-Net is not available for URBAN3D dataset. To address this, we use an adversarial network to discriminate between the super-resolved outputs generated by SR-Net and real high-resolution imagery. The discriminator network(\textbf{D}) is trained on high resolution tiles from ISPRS data and outputs of SR-Net on URBAN3D samples. The predictions of \textbf{D} on outputs of SR-Net are used to compute the adversarial loss. Adversarial loss for super-resolution was proposed for SR-GAN~\cite{ledig2017photo} in conjunction with MSE loss for supervision. We use adversarial loss along with a stage-wise mix of L2 and L1 losses instead.\\
\textbf{Loss functions}
Besides the adversarial loss from the discriminator network, we use binary cross entropy (BCE) criterion to supervise SUNET for the building footprint segmentation task. SR-Net is trained using adversarial loss as well as direct supervision on source domain samples. In the initial 1000 epochs of training we use L2 loss on SR-Net output. L1 loss is used in subsequent epochs. L1 preserves colors and luminance. An error is weighed equally regardless of the local structure. Following the trends reported in~\cite{zhao2017loss}, we opt for using L2 loss initially to ensure the generator network (SR-Net) learns slowly enough to not outpace the discriminator. After that, we opt for L1 which is reported to provide a steeper descent in loss curves~\cite{zhao2017loss}.
\par\textbf{ Material label transfer} Our objective is to harvest the rich and highly detailed semantic/material annotations from the high resolution dataset to make inference on the super-resolved URBAN3D images. ISPRS annotations cover 6 semantic classes. URBAN3D on the other hand provides building footprint labels only. Material labels are harder to obtain than semantic labels. Moreover, in low resolution satellite imagery, it is difficult even for human annotators to reliably predict material labels of urban structures (e.g. building roofs, roads, pavements etc.). Therefore, direct annotation of large quantities of training data is challenging. An alternative solution to this problem is to obtain smaller quantities of urban material labels on high resolution datasets. UDA techniques can thereafter be used for automatic label transfer from source to target domain. Super-resolution can increase the fidelity of such transfer by bridging the domain gap and enhancing certain texture information. It can also help with enhanced detection of outlier regions (small objects on roof, shadows etc.)
\begin{figure}[thb]
    \centering
    \includegraphics[scale=0.4]{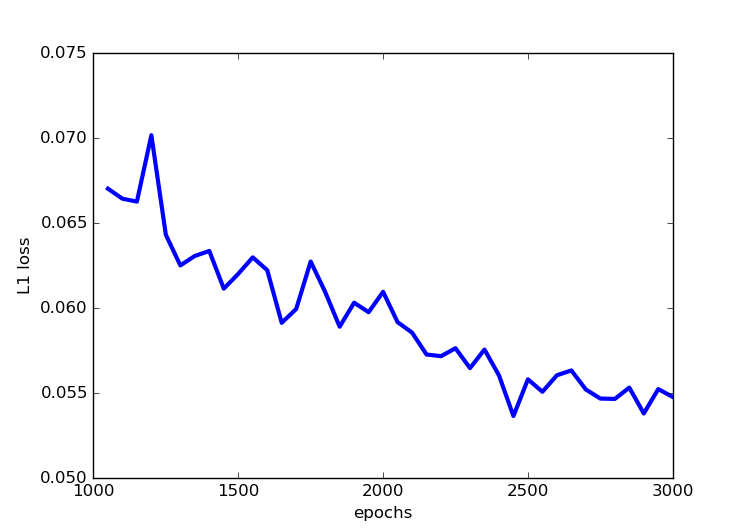}
    \caption{L1 loss computed on output of SR-Net for training samples from ISPRS dataset}\label{PlotL1}
\end{figure}
\section{Experiments}\label{sec:Expt}
We augment the Stacked U-Net (SUNET) architecture used for land cover classification \cite{ghosh2018stacked} for unsupervised satellite
imagery super-resolution using adversarial learning. We initialize the SUNET with pre-trained weights from the DeepGlobe land cover classification challenge. We jointly optimize the network on 512x512 training samples from two datasets.
\par \textbf{Dataset A:  URBAN3D} Samples consist of ortho-rectified RGB images at 50 cm ground sample distance. Labels are building footprints and background.
\par \textbf{Dataset B:  ISPRS}  Potsdam 2D Semantic labeling contest samples are RGB orthophotos at 5cm ground sample distance. The dataset contains 38 patches of same size. Labels consist of 6 semantic classes - building, cars, trees, low vegetation, roads(Impervious surfaces), clutter(background).
\par
\begin{figure*}[tbh]
    \centering
    \includegraphics[scale=0.56]{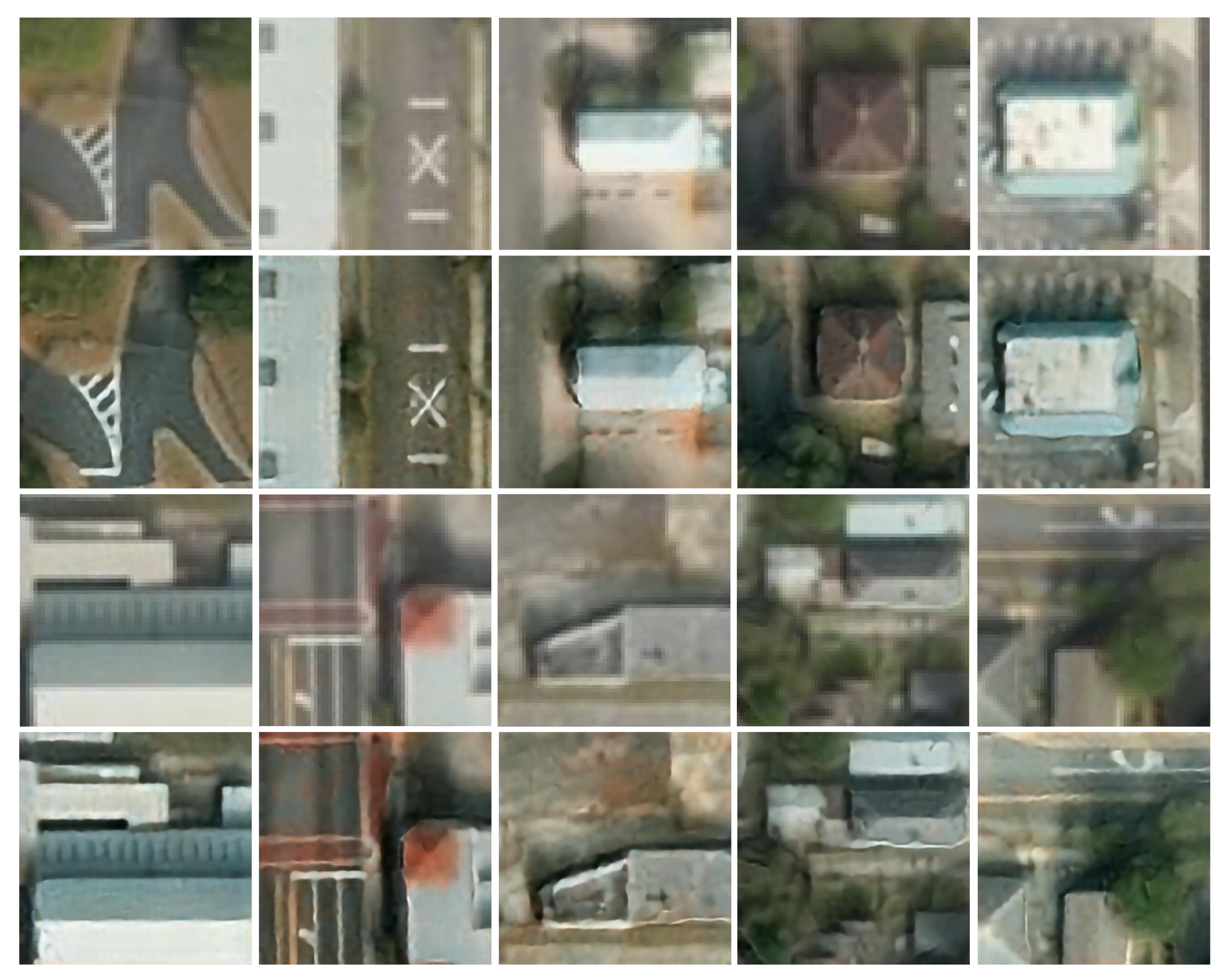}
    \caption{Super-resolved remote sensing imagery from URBAN3D dataset. First and third row consist of original low resolution 64x64 tiles whereas second and fourth contain their super-resolved 512x512 versions.}\label{Fig:Res}
\end{figure*}

 For training the networks, we sample 64x64 tiles from images in A and scale them to 512x512 using bi-linear interpolation (BLI). We crop 512x512 tiles from samples in B and introduce Gaussian blur. SUNET is jointly optimized using binary cross entropy (BCE) loss on samples from both A and B to perform building footprint segmentation. The features of SUNET are thus aligned for both datasets and further used for super-resolution using 4 pixel-shuffling layers in SR-Net. SR-Net layers are initialized using ~\cite{he2015delving} and supervised using high resolution ground truth tiles from source domain (ISPRS). We apply an adversarial loss on outputs (for samples of A and B) using a discriminator network D. D is trained to discriminate between super-resolved outputs on A and high resolution ground truth tiles from B. Network parameters are optimized using the RMSProp algorithm~\cite{tieleman2012lecture}. The models are implemented in PyTorch.
Figure~\ref{PlotL1} shows drop in  L1 loss during training on ISPRS samples. Computing L1 loss in target domain (URBAN3D) is not possible because of lack of high resolution data. Figure~\ref{Fig:Res} shows super-resolved outputs on some tiles from the URBAN3D data.
\section{Discussion}\label{sec:disc}
\begin{figure}
    \centering
    \includegraphics[scale=0.3]{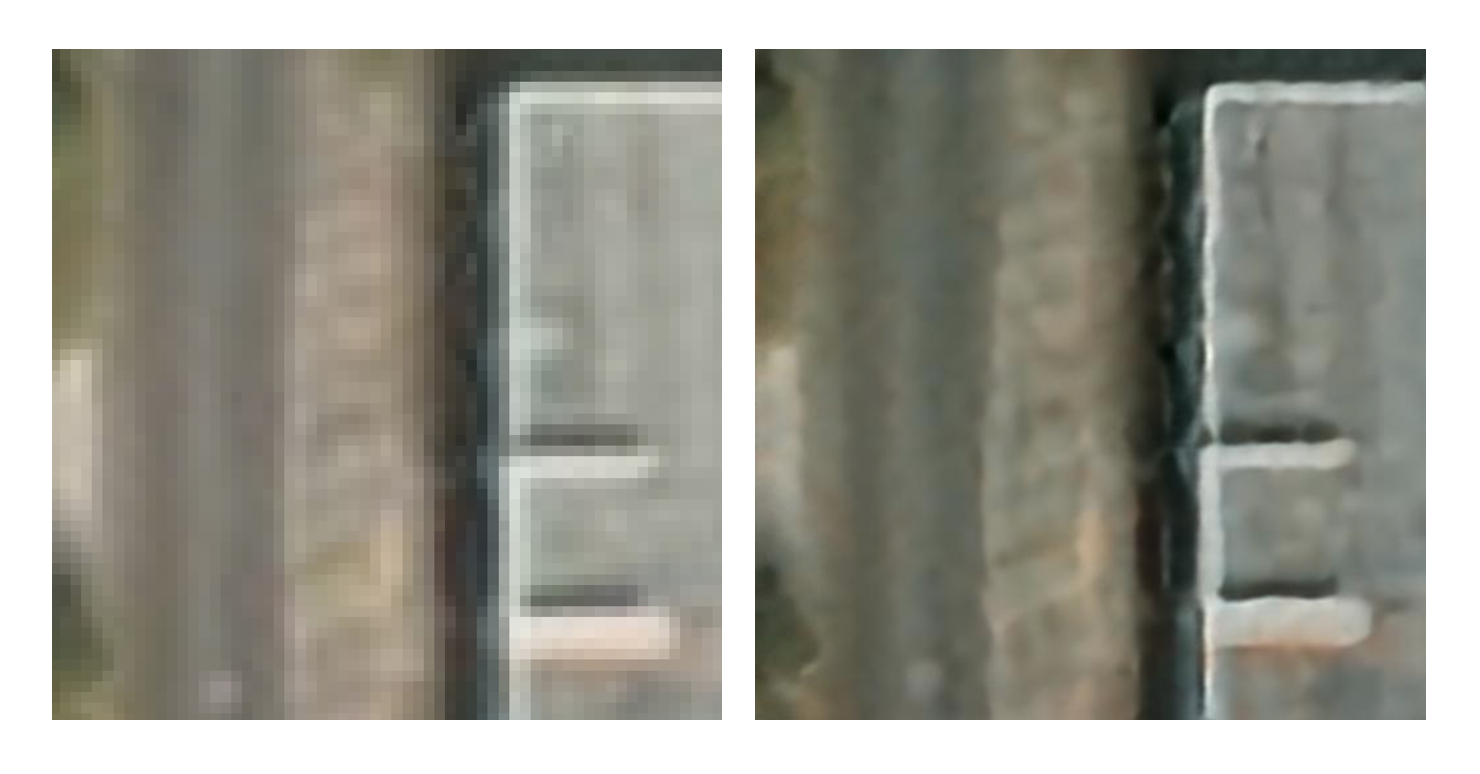}
    \caption{Super resolution results in more clear boundaries between different material types on a roof}\label{roofstruct}
\end{figure}
Compared to URBAN3D data, ISPRS images have ten times higher resolution. The segmentation network (SUNET) is provided with 512x512 upscaled inputs generated from 64x64 tiles randomly cropped from URBAN3D samples. The ISPRS input samples in the mixed training batches are obtained by randomly cropping 512x512 tiles. Thus the inputs from the two domains are at slightly different zoom levels.  Moreover, in order to simulate lower resolution inputs corresponding to ISPRS images, we introduce a Gaussian blur. URBAN3D dataset consists of images from Jacksonville, FL.  Different lighting, abundance(or lack) of water bodies and different nature of vegetation at Potsdam and Jacksonville contribute to the domain gap between these two datasets. The proposed approach overcomes this gap and learns to improve the resolution of URBAN3D images. Figure~\ref{Fig:Res} shows several examples where boundaries of tiny buildings are much sharper in the output, road markings are clearer and details such as small structures (chimneys), roof shape, shadows are enhanced. Figure ~\ref{roofstruct} provides a specific example where boundaries between different materials (structures on a roof) are enhanced. \vspace{-10pt}
\section{Acknowledgements}\vspace{-8pt}
The research is based upon work supported by the Office of the Director of National Intelligence (ODNI), Intelligence Advanced Research Projects Activity (IARPA), via DOI/IBC Contract Number D17PC00287. The views and conclusions contained herein are those of the authors and should not be interpreted as necessarily representing the official policies or endorsements, either expressed or implied, of the ODNI, IARPA, or the U.S. Government. The U.S. Government is authorized to reproduce and dis- tribute reprints for Governmental purposes notwithstanding any copyright annotation thereon.

\bibliographystyle{IEEEtran}
\bibliography{egbib}

\end{document}